%% file: acl_latex.tex
\documentclass[11pt]{article}

\usepackage[preprint]{acl}
\usepackage{times}
\usepackage{latexsym}
\usepackage[T1]{fontenc}
\usepackage[utf8]{inputenc}
\usepackage{microtype}
\usepackage{inconsolata}
\usepackage{graphicx}

\usepackage{hyperref}
\usepackage{booktabs}
\usepackage{amssymb}
\usepackage{amsmath}

\newcommand{\method}{DIAL-S2ST}

\title{\method: \\ Learning Latency Control for Simultaneous Speech-to-Speech Translation}

\author{Jeongsoo Choi \quad Junseok Ahn \quad Joon Son Chung\\
KAIST\\
{\small\texttt{\{jeongsoo.choi, junseok.ahn, joonson\}@kaist.ac.kr}}}

\begin{document}
\maketitle

\begin{abstract}
We propose \method{}, a post-training method for learning inference-time latency control in simultaneous speech-to-speech translation models.
Our approach reuses pretrained translation capabilities through self-distillation from lookahead predictions, while limiting the influence of large text distillation losses to account for the imitation gap.
This provides supervision for earlier emission without paired target data or reward computation.
To make the learned timing adjustable at inference, we jointly train a single LoRA adapter across scales with matched lookahead guidance.
Experiments on French-to-English translation with Hibiki demonstrate that only 10 hours of source speech enable this adaptation, providing flexible latency control with a favorable quality-latency trade-off.
\end{abstract}

\section{Introduction}
Speech-to-speech translation enables communication and access to spoken content across languages~\cite{jia2019direct, kim2024textless, seamless2025joint}.
Waiting for a complete source utterance before translating, however, introduces delays that limit its use in real-time settings~\cite{dugan2023learning}.
Simultaneous speech-to-speech translation (SimulS2ST) addresses this limitation by producing translated speech while the source utterance is still unfolding~\cite{ma2021direct}.
To do so, a SimulS2ST model must continually decide whether to read more source speech or to write translated speech.
Reading more provides additional context for translation but delays the output, so SimulS2ST involves a trade-off between translation quality and latency~\cite{papi2022does}.
Since some applications prioritize quality while others require low latency, the quality-latency balance should be adjustable at inference time~\cite{zhang2024streamspeech}.

\begin{figure*}[t]
\begin{center}
\centerline{\includegraphics[width=0.8\textwidth]{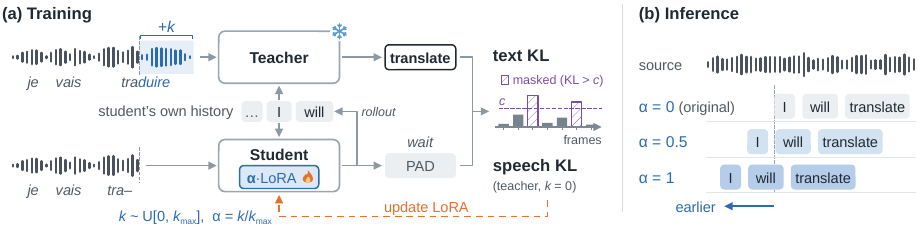}}
\vspace{-1mm}
\caption{Overview of \method{} training and inference.}
\label{fig:DIAL}
\vspace{-8mm}
\end{center}
\end{figure*}

SimulS2ST research has developed explicit mechanisms for adjusting latency at inference time without retraining.
StreamSpeech~\citep{zhang2024streamspeech} controls latency through the input chunk size, while Seamless~\citep{barrault2023seamless} adjusts a threshold for emitting translated text.
Explicit latency control is less straightforward in multistream models such as Hibiki~\citep{labiausse2025high} and Hibiki-Zero~\citep{labiausse2026simultaneous}.
These models jointly generate translated text and speech as parallel token streams~\citep{defossez2024moshi}.
The generated streams encode both translation content and its timing relative to the source, coupling when to speak with what to say.
Their latency is thus determined by generation behavior learned during training.
Once trained, neither model has a separate read/write policy to adjust at inference time.

Without such a policy, a simple way to adjust latency at inference is to bias the logits of special tokens that mark when to speak, as previously suggested~\cite{arora2025talking}. 
Our experiments, however, show that this approach yields an unfavorable quality-latency trade-off.
Another approach is supervised fine-tuning on temporally shifted target alignments~\cite{labiausse2025high}, but this adapts the model to a single prescribed timing, requiring separate training for each operating point.
The challenge is therefore to enable inference-time latency control within a single model while maintaining a favorable trade-off.

To this end, we propose \method{} (\textbf{D}istillation for \textbf{I}nference-time \textbf{A}djustable \textbf{L}atency), a self-distillation method for learning inference-time latency control in pretrained multistream SimulS2ST models. 
Following on-policy self-distillation~\cite{zhao2026self}, we use the frozen pretrained model with source lookahead as a teacher and adapt the same model without lookahead as a student to match the teacher’s text predictions on student-generated trajectories.
Matching a teacher that has already heard upcoming source speech encourages the student to emit translation content earlier, requiring only source speech without paired targets or reward computation.
However, the same lookahead introduces an imitation gap~\cite{weihs2021bridging} because some teacher predictions rely on source speech the student has not yet heard.
We therefore mask exceptionally large text distillation losses to limit their influence.
We train a single LoRA adapter~\cite{hu2022lora} for the student by pairing larger scales with longer lookaheads, enabling latency control via its scale at inference.

We evaluate \method{} on French-to-English translation using Hibiki.
With approximately 10 hours of source speech and 500 optimization steps, \method{} achieves a more favorable quality-latency trade-off than inference-time logit biasing.
A single adapter supports a broad range of latency operating points at inference.
These results show that brief source-only adaptation can improve the quality-latency trade-off while providing inference-time control over the learned emission timing.

\section{Method}
\label{sec:method}
\subsection{Multistream speech-to-speech translation}
\label{sec:background}
We consider multistream SimulS2ST models that represent source speech, target text, and target speech as parallel, time-synchronous token streams~\cite{labiausse2025high,labiausse2026simultaneous}.
The causal Mimi codec~\cite{defossez2024moshi} encodes source and target waveforms $X$ and $Y$ into discrete tokens $A^X_{t,q}$ and $A^Y_{t,q}$ at $f_r=12.5$~Hz, where $t$ and $q$ denote the time step and codebook indices, respectively, and $Q$ is the number of codebooks.

A Temporal Transformer reads the available source tokens and generated history to predict a target text token $W_t$ at each step.
Its output and $W_t$ then condition a Depth Transformer that autoregressively generates target speech tokens across codebooks.
All streams run at $f_r$, with acoustic tokens $A^Y_{t,\geq2}$ delayed by two frames relative to the semantic tokens $A^Y_{t,1}$.
Following Hibiki~\cite{labiausse2025high}, we index the audio streams after applying this delay, so $A^Y_{t,\ge 2}$ denotes the acoustic tokens generated at step $t$.

The target text stream is aligned with the target speech at the word level. 
Since text tokens typically occupy fewer frames than the corresponding speech, \texttt{PAD} fills the remaining frames until the next word, acting as a wait token in the text stream.
Translation content and emission timing are therefore learned through the same token predictions.

\subsection{Lookahead on-policy self-distillation}
\label{sec:lookahead}
Our central idea is to let the pretrained model teach itself to speak earlier by imitating its own predictions made with additional source context (Fig.~\ref{fig:DIAL}~(a)).
We use the frozen pretrained model $p_{\theta_0}$ as teacher $p_T$, while the causal student $p_{S,\alpha}$ shares its frozen base parameters and adds a trainable LoRA~\cite{hu2022lora} adapter $\phi$ to the Temporal Transformer, with its update scaled by $\alpha$. 
Following on-policy distillation~\cite{agarwal2024policy,zhao2026self}, we sample a text-and-speech trajectory $h\sim p_{S,\alpha}(\cdot\mid A^X)$ for each utterance $X$ in a source-only dataset $\mathcal{D}_{\mathrm{src}}$ and condition both models on its prefix at each step, leaving this prefix implicit in Eqs.~\eqref{eq:text-distillation-loss}-\eqref{eq:speech-distillation-loss}. 
This exposes the student to its own generated histories, as during inference.

\paragraph{Shifting text earlier while anchoring speech.}
We distill the teacher's text predictions with a lookahead of $k$ seconds into the 
causal student:
\begin{equation}
d_t^{\mathrm{txt}}
= D_{\mathrm{KL}}\Bigl(
p_T^{\mathrm{txt}}(\cdot\mid A^X_{\leq t+f_r k})
\Big\|\,
p_{S,\alpha}^{\mathrm{txt}}(\cdot\mid A^X_{\leq t})
\Bigr).
\label{eq:text-distillation-loss}
\end{equation}
With this additional context, the teacher can already emit words for which the model without lookahead would still wait.
Matching these predictions therefore encourages the student to emit translation content earlier from the source speech heard so far.

Since the adapter also changes the Temporal Transformer output that conditions the Depth Transformer, adapting text timing can alter the generated speech. 
To mitigate this, we regularize the student’s speech predictions toward those of the teacher without lookahead, given the same source context and text.
The codebook-level distillation loss is
\begin{equation}
d_t^{(q)}
= D_{\mathrm{KL}}\Bigl(
p_T^{(q)}(\cdot\mid A^X_{\leq t})
\Big\|\,
p_{S,\alpha}^{(q)}(\cdot\mid A^X_{\leq t})
\Bigr),
\label{eq:speech-distillation-loss}
\end{equation}
where both models condition on the student’s text token $W_t$ and tokens from preceding codebooks $A^Y_{t,<q}$.
We weight the acoustic losses by $\lambda$.

\paragraph{Masking large text KL.}
Source lookahead can create an imitation gap because some teacher predictions rely on source speech the student has not yet heard.
To limit the contribution of large teacher-student discrepancies, we mask text KL terms above a fixed threshold $c$. 

\noindent The frame-level loss is
\begin{equation}
\ell_t
=d_t^{\mathrm{txt}}\,
\mathbf{1}[d_t^{\mathrm{txt}}\leq c]
+d_t^{(1)}
+\lambda\sum_{q=2}^{Q}d_t^{(q)}.
\label{eq:frame-loss}
\end{equation}
The total distillation loss is
\begin{equation}
\mathcal{L}(\phi)
=\mathbb{E}_{X,k,h}
\left[\frac{1}{T}\sum_{t=1}^{T}\ell_t\right],
\label{eq:distillation-objective}
\end{equation}
where $T$ is the length of $h$, and $k$ and $\alpha$ are set as in Eq.~\eqref{eq:lookahead-scale}.

\subsection{Latency control with a single adapter}
\label{sec:latency-control}
For each training sequence, we sample a lookahead $k$ and set the corresponding adapter scale:
\begin{equation}
k\sim\mathcal{U}[0,k_{\max}],
\qquad
\alpha(k)=\frac{k}{k_{\max}},
\label{eq:lookahead-scale}
\end{equation}
where $k_{\max}$ is the maximum lookahead.
Unlike training at a single scale, this pairing explicitly teaches how emission timing should vary with the adapter scale.
At inference, we control latency by adjusting $\alpha$ alone.
A larger $\alpha$ yields earlier output (Fig.~\ref{fig:DIAL}~(b)), and setting $\alpha=0$ exactly recovers the pretrained model.

\section{Experimental Setup}
\label{sec:experiments}
\subsection{Datasets}
We evaluate all models on the French-to-English test set of CVSS~\cite{jia2022cvss}, a widely used speech-to-speech translation dataset. 
CVSS is built on CoVoST~2~\cite{wang2021covost}, which uses source speech from Common Voice~\cite{ardila2020common}. 
Unless otherwise stated, post-training uses approximately 10 hours of French source speech from the out-of-domain FLEURS~\cite{conneau2022fleurs} training split.
We also conduct in-domain post-training on a random subset of the CVSS training split with comparable total duration.

\subsection{Evaluation metrics}
Translation quality is measured by ASR-BLEU, computed with SacreBLEU~\cite{post2018call} between Whisper-medium~\cite{radford2023robust} transcripts of generated speech and reference translations.
Latency is measured by LAAL~\cite{papi2022over} and End Offset~\cite{labiausse2025high}, capturing average translation delay and completion delay, respectively.
Voice transfer is measured by the cosine similarity between speaker embeddings extracted from source and generated speech using WavLM-Large~\cite{chen2022wavlm} fine-tuned for speaker verification, and naturalness is estimated with UTMOS~\cite{saeki2022utmos}.

\subsection{Baselines}
We compare \method{} with four open-source SimulS2ST models.
StreamSpeech~\cite{zhang2024streamspeech} is an encoder-decoder model trained through multi-task learning for speech recognition, translation, and synthesis.
Its inference-time latency is adjusted by varying the input chunk size.
Seamless~\cite{barrault2023seamless} is a multilingual encoder-decoder model that uses monotonic attention to decide when to emit translated text, with its decision threshold controlling inference-time latency.
Hibiki~\cite{labiausse2025high} is a decoder-only multistream model that learns emission timing from word-level aligned training data and serves as our base model.
Hibiki-Zero~\cite{labiausse2026simultaneous} builds on this architecture and optimizes the quality-latency trade-off during training through reinforcement learning.

\subsection{Implementation details}
The LoRA adapter has rank 16 and is applied to every linear layer of the Temporal Transformer, resulting in 16\,M trainable parameters.
We train for 500 steps with 16 rollouts per step using AdamW optimizer~\cite{loshchilov2019decoupled}, a learning rate of $5\times10^{-5}$, and linear warm-up.
For distillation, we set the acoustic loss weight $\lambda$ to 0.02, the text KL masking threshold $c$ to 10 nats, and $k_{\max}$ to 2\,s.
Every configuration is trained with three random seeds, and we average the three adapters using uniform model soups~\cite{wortsman2022model}.

\section{Experimental Results}
\label{sec:results}

\begin{figure}[t]
\begin{center}
\centerline{\includegraphics[width=\columnwidth]{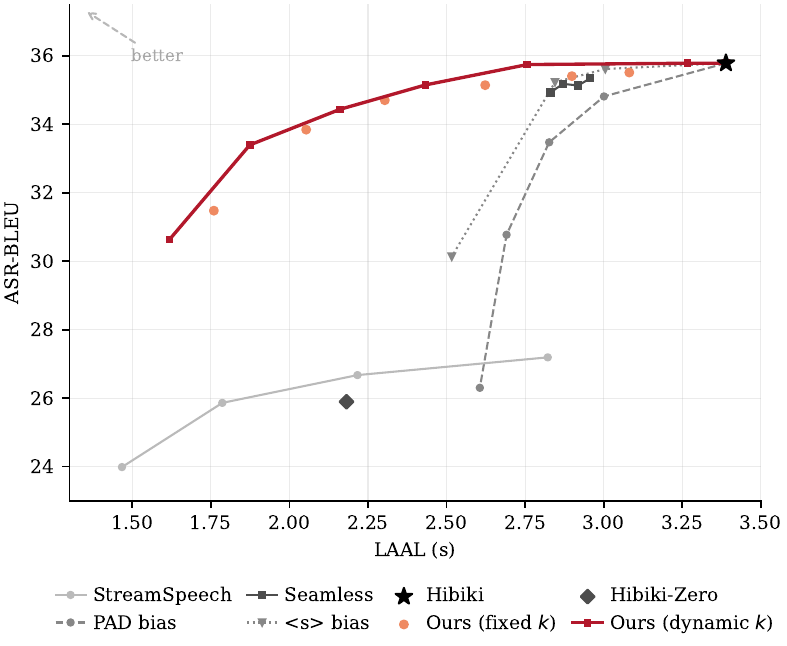}}
\vspace{-3mm}
\caption{
    Quality-latency trade-off on the CVSS Fr$\to$En test set.
    Connected points represent different inference-time latency settings of the same model, while isolated points correspond to distinct models.
}
\label{fig:tradeoff}
\vspace{-10mm}
\end{center}
\end{figure}

\subsection{Main results}
\label{sec:main}
Fig.~\ref{fig:tradeoff} compares the quality-latency trade-offs of \method{} and the baseline methods.
For StreamSpeech, latency is adjusted using input chunk sizes ranging from 320 to 2560\,ms.
\method{} achieves substantially higher translation quality across a broad range of comparable latencies.
For Seamless, varying the decision threshold from 0.4 to 0.7 provides a limited range of latency adjustment.

We also evaluate inference-time logit biasing on Hibiki, the base model of \method{}.
Applying biases of $-1$, $-2$, $-3$, and $-4$ to the \texttt{PAD} logit reduces latency but leads to a sharp decline in translation quality.
A similar trend is observed when biasing the \texttt{<s>} (BOS) logit by $+6$, $+10$, and $+14$ to encourage an earlier start to translation.
Increasing this bias further causes the model to generate speech without waiting for source input.
\method{} enables latency adjustment over a broad range at inference while achieving a competitive Pareto frontier with a single adapter.

\subsection{Ablations}
\label{sec:ablation}
\paragraph{Post-training strategy.}
Table~\ref{tab:main} compares \method{} with two alternatives at matched latency.
For self-training, we first generate translated speech from the source speech using the pretrained model.
We then shift the generated speech earlier relative to the source and use it as the target for supervised fine-tuning.
The off-policy variant replaces student rollouts with lookahead teacher rollouts while retaining the same distillation objective.
\method{} achieves the highest ASR-BLEU at all operating points, with its advantage over off-policy distillation increasing at lower latency.
These results support distilling on student-generated trajectories when adapting the model to emit translations earlier.

\input{tables/main}

\input{tables/ablations}

\paragraph{Design choices.}
Table~\ref{tab:ablation} shows that masking large text KL terms improves translation quality at similar latency.
Removing acoustic KL lowers speaker similarity, and removing both semantic and acoustic KL further degrades translation quality and voice transfer.
Grouped text KL, adapted from ASPIRin~\cite{hsiao2026aspirin}, aggregates text probabilities into groups for padding, word starts, and all remaining tokens.
The full text distribution achieves a comparable or better quality-latency trade-off, showing that \method{} can learn latency control without explicitly separating timing decisions from lexical predictions.
Replacing forward KL with reverse KL in all distillation terms causes silence collapse.
Finally, out-of-domain FLEURS training yields performance comparable to in-domain CVSS training, supporting adaptation with source speech from outside the evaluation domain.

\section{Conclusion}
\label{sec:conclusion}
In this paper, we introduced \method{} for learning latency control in pretrained multistream SimulS2ST models.
We explored how lookahead self-distillation can reuse pretrained translation capabilities to adapt emission timing while accounting for the imitation gap.
Learning this adaptation jointly across adapter scales allows the resulting emission behavior to be controlled through a single adapter.
Experiments on French-to-English translation with Hibiki demonstrate that only a small amount of source speech is sufficient to make the model’s learned emission timing adjustable at inference.

\section*{Limitations}
Our experiments focus on the French-to-English Hibiki model and primarily use utterance-level speech.
Latency is controlled by adjusting the adapter scale, with the resulting latency reduction varying across utterances.
The relationship between adapter scale and measured latency is characterized empirically for the evaluated model and dataset.
Translation quality and latency measurements rely on ASR transcriptions and forced word alignments, respectively, while speaker similarity and speech naturalness are assessed using automatic metrics.
Future work could extend the study to other pretrained models and language pairs, examine latency control in long-form translation, and include human evaluation of translation adequacy and speech naturalness.

\section*{Ethical Considerations}
This work uses publicly available datasets and pretrained models, including FLEURS, CVSS, and Hibiki.
Our experiments do not involve additional human-subject data collection or user studies.
In practical deployment, generated speech may contain translation errors or be misused for voice impersonation.
We therefore encourage responsible use that complies with resource licenses, respects speaker consent, clearly discloses machine-generated translations, and includes appropriate human oversight in sensitive settings.

\bibliography{custom}

\end{document}

%% file: tables/main.tex
\begin{table}[t]
\begin{center}
\begin{footnotesize}
\setlength{\tabcolsep}{4pt}
\resizebox{0.7\columnwidth}{!}{
\begin{tabular}{@{}lccc@{}}
\toprule
& \multicolumn{3}{c}{LAAL (s)} \\
\cmidrule(l){2-4}
& 1.8 & 2.3 & 2.8 \\
\midrule
Ours                                        & \textbf{32.5} & \textbf{34.8} & \textbf{35.6} \\
\quad w/ self-training (shifted rollouts)   & 32.0 & 33.2 & 34.3 \\
\quad w/ off-policy (teacher rollouts)      & 31.1 & 34.6 & 35.5 \\
\bottomrule
\end{tabular}}
\end{footnotesize}
\end{center}
\vspace{-3mm}
\caption{ASR-BLEU at matched LAAL on the CVSS Fr$\to$En test set.}
\label{tab:main}
\vspace{-3mm}
\end{table}

%% file: tables/ablations.tex
\begin{table}[t]
\begin{center}
\begin{footnotesize}
\setlength{\tabcolsep}{4pt}
\resizebox{\columnwidth}{!}{
\begin{tabular}{@{}lccccc@{}}
\toprule
& ASR-BLEU & LAAL & End & Speaker & UTMOS \\
& $\uparrow$ & $\downarrow$ & Offset $\downarrow$ & Sim. $\uparrow$ & $\uparrow$ \\
\midrule
Ours                        & 34.4 & 2.16 & 1.73 & 0.391 & 3.41 \\
\quad w/o text KL masking   & 33.1 & 2.20 & 1.73 & 0.390 & 3.40 \\
\quad w/o acoustic KL       & 34.1 & 2.08 & 1.75 & 0.385 & 3.41 \\
\quad w/o speech KL         & 32.7 & 2.16 & 1.69 & 0.337 & 3.33 \\
\quad w/ grouped text KL    & 34.2 & 2.22 & 1.94 & 0.392 & 3.42 \\
\quad w/ reverse KL         & \multicolumn{5}{c}{\textit{collapses to silence}} \\
\quad w/ CVSS training data & 34.0 & 2.06 & 1.51 & 0.387 & 3.42 \\
\bottomrule
\end{tabular}}
\end{footnotesize}
\end{center}
\vspace{-3mm}
\caption{Ablations of \method{} ($\alpha=0.5$).}
\label{tab:ablation}
\vspace{-5mm}
\end{table}